\documentclass{article} 
\usepackage[preprint]{colm2026_conference}

\usepackage[T1]{fontenc}
\usepackage[utf8]{inputenc}
\usepackage{amsmath}
\usepackage{amssymb}
\usepackage{mathtools}
\usepackage{microtype}
\usepackage{graphicx}
\usepackage{booktabs}
\usepackage{algorithm}
\usepackage{algpseudocode}
\usepackage{hyperref}
\usepackage{url}
\usepackage{lineno}
\usepackage{enumitem}

\definecolor{darkblue}{rgb}{0, 0, 0.5}
\hypersetup{colorlinks=true, citecolor=darkblue, linkcolor=darkblue, urlcolor=darkblue}

\title{SLIM-RL: Risk-Budgeted Random-Masking RL \\ for Diffusion LLMs Without Trajectory Slicing}

\author{Ruikang Zhao \thanks{Correspondence to: Ruikang Zhao (\texttt{ruikangzhao@gmail.com}) and Ligong Han (\texttt{hanligong@gmail.com}).} \\ 
Technical University of Denmark \\
\And
Zhenting Wang \\
MBZUAI Institute of Foundation Models \\
\And
Han Gao \\
Iowa State University \\
\And
Ligong Han \textsuperscript{*}  \\
Red Hat AI Innovation \& MIT--IBM Watson AI Lab }

\begin{document}

\ifcolmsubmission
\linenumbers
\fi

\maketitle
\begin{abstract}
    Reinforcement learning for diffusion large language models (dLLMs) has largely moved to trajectory-aware methods. The current state of the art, TraceRL, holds that random masking is mismatched with the model's inference trajectory, and it reconstructs that trajectory during training by slicing each rollout into up to $\lceil K/s\rceil$ trajectory-aligned training samples, a cost that grows with the block size $K$. We show that this mismatch can be mitigated without reconstructing the trajectory. Our method, \textbf{SLIM-RL}, bounds the commit risk of each rollout step with a $\tau$-budget decoder, reducing aggregate commit risk in the training data. During optimization, SLIM-RL trains on these risk-controlled rollouts with a trace-free random-masking objective that adapts variance-reduction tools, combining sequence-level importance sampling, deterministic quadrature over masking levels under a mean-preserving, monotonically decreasing per-block mask schedule that we introduce. On SDAR-4B, SLIM-RL matches TraceRL's best MATH500 accuracy on only $0.46\times$ its training samples at block size 16, improving over TraceRL by $6.32\%$ on MATH500 and $11.05\%$ on GSM8K under matched dynamic sampling. At block size 4, the $4$B SLIM-RL surpasses the larger LLaDA-8B and Dream-7B dLLMs on math, exceeding LLaDA-8B by $10.76\%$ on MATH500 while staying below the autoregressive Qwen2.5-7B. On code, it improves over TraceRL by $4.20\%$ on MBPP and $3.65\%$ on HumanEval. The $\tau$-budget decoder transfers training-free across LLaDA, Dream, and SDAR. The source code is available at https://github.com/laolaorkkkkk/SLIM-RL . 
\end{abstract}

\section{Introduction}
\label{sec:Introduction}
Diffusion Large Language Models (dLLMs)~\citep{nie2026large,sahoo2024simple,ye2025dream} generate text by iteratively denoising a masked sequence, refining many positions at once instead of strictly left to right. Block-wise variants~\citep{arriola2025block,cheng2026sdar} decode consecutive blocks autoregressively while denoising within each block, which restores KV-cache reuse. Reinforcement learning has become the standard way to improve their reasoning~\citep{zhao2026d1,wang2025revolutionizing,hu2026lightningrl,he2025mdpo,zhu2025dirl,tang2026wd1weightedpolicyoptimization,liu2026stp,zhu2025llada15variancereducedpreference}.

Early work applies RL to randomly masked targets~\citep{zhao2026d1,yang2026mmada}, but random masking yields high-variance gradient estimates~\citep{zhu2025llada15variancereducedpreference}, and recent work has therefore moved toward trajectory-aware methods~\citep{huang2026reinforcing,wang2026d2improvingreasoningdiffusion}. The current state of the art, TraceRL~\citep{wang2025revolutionizing}, trains not on random masks but on the model's exact decoding trajectory, on the view that under random masking the post-training objective is mismatched with the trajectory the model follows at inference.

Preserving the trajectory is not free. To train on each decoding step in order, TraceRL slices a single rollout into up to $\lceil K/s\rceil$ trajectory-aligned training samples, one forward each, where $K$ is the block size and $s{\ge}1$ a shrinkage factor~\citep{wang2025revolutionizing}. At full fidelity ($s{=}1$) one rollout becomes up to $K$ samples, so the data grows with $K$. Raising the shrinkage $s$ bounds the cost but aggregates $s$ consecutive decoding steps into one slice and discards their internal order, trading trajectory fidelity for cost. A larger block therefore makes a trade-off between slicing cost and trajectory fidelity, which we quantify in Section~\ref{sec:exp_efficiency}.

The question is whether reconstructing the exact trajectory is necessary. Without slicing, a dLLM-RL run still faces two weaknesses at two independent stages, rollout generation and optimization. Rollouts are generated with dynamic sampling~\citep{wu2025fast,yu2025dimple}, the common rollout decoder across dLLM-RL methods including TraceRL, which commits every token whose confidence exceeds a fixed threshold $\tau$. Because this rule is pointwise, a step that unmasks many positions only marginally above $\tau$ over-commits and injects several errors into the rollout. At optimization, random masking makes the policy-gradient estimate high-variance~\citep{zhu2025llada15variancereducedpreference}.

In this work, we address both stages and call the resulting recipe \textbf{SLIM-RL}, which combines three components developed in Section~\ref{sec:method}. A $\tau$-budget dynamic-unmasking decoder, a single-pass training-free rule shared by rollout and inference, caps each step's cumulative confidence-based uncertainty. A variance-reduced framework then updates the policy, built from a sequence-level length-normalized ratio~\citep{zheng2025group}, deterministic quadrature over masking levels~\citep{rojas2025improving}, and an unnormalized advantage~\citep{liu2025understanding}. A mean-preserving, monotonically decreasing per-block mask schedule front-loads the policy-gradient signal onto the earliest, most-conditioned block.

On SDAR-4B, SLIM-RL outperforms trajectory-aware TraceRL on math at block size 16 and on both math and code at block size 4, where the 4B model also surpasses the larger LLaDA-8B and Dream-7B diffusion models, and reaches TraceRL's best accuracy on under half the training data (Table~\ref{tab:dataeff}). The margin widens with block size, and Section~\ref{sec:exp} reports the per-benchmark results.

Our contributions are summarized as follows:

\begin{itemize}

    \item We show that reconstructing the exact decoding trajectory is not required to match trajectory-aware RL at equal training cost. With rollout commit risk bounded and the masking objective variance-reduced, trace-free random masking is on par with trajectory-aligned slicing at block size 4 and outperforms it at block size 16.

    \item We introduce a $\tau$-budget dynamic-unmasking decoder that commits only the largest low-uncertainty subset of positions whose cumulative uncertainty $\sum_i(1-p_i)$ stays within a calibrated budget $m(1-\tau)$, in contrast to dynamic sampling's per-position threshold. It is training-free and transfers across LLaDA, Dream, and SDAR.

    \item We introduce a mean-preserving, monotonically decreasing per-block mask schedule that concentrates the policy-gradient signal on the earliest, most-conditioned block, adding $5.91\%$ on MATH500 over random masking.

    \item On SDAR-4B at block size 16, SLIM-RL achieves $6.32\%$ and $11.05\%$ higher accuracy than TraceRL on MATH500 and GSM8K under matched decoding, and reaches TraceRL's best accuracy on $0.46\times$ the training data. At block size 4, it achieves higher accuracy than the larger LLaDA-8B and Dream-7B diffusion models across math and code, and improves over TraceRL on code by $4.20\%$ on MBPP and $3.65\%$ on HumanEval. On SDAR-1.7B, SLIM-RL reaches TraceRL's best on $0.76\times$ the training data.


\end{itemize}

\section{Preliminaries}
\label{sec:prelim}

\subsection{Diffusion Large Language Models}
\label{sec:prelim_dllm}

\paragraph{Forward and reverse processes.}
Given a prompt $x$ and a clean response $y=(y^{1},\ldots,y^{L})$, a diffusion large language model (dLLM)~\citep{nie2026large,ye2025dream} defines a forward corruption process and a learned reverse generation process. The forward process independently replaces each token with the mask symbol $[\mathrm{MASK}]$ at masking level $t\in[0,1]$:
\begin{equation}
    q_{t}(y_{t}\mid y)
    =
    \prod_{i=1}^{L}
    q_{t}(y_{t}^{i}\mid y^{i}),
    \label{eq:forward}
\end{equation}
where each token is independently masked according to $q_{t}(y_{t}^{i}{=}[\mathrm{MASK}]\mid y^{i})=t$ and $q_{t}(y_{t}^{i}{=}y^{i}\mid y^{i})=1-t$, so that $y_{0}$ is the clean response and $y_{1}$ is fully masked. For reverse process, starting from $y_{1}$, the model iteratively predicts masked positions and resamples a less-noised $y_{r}$ from $y_{t}$ ($r<t$) until $y_{0}$ is recovered.

\paragraph{Block-wise dLLMs.}
Vanilla dLLMs denoise the full response under bidirectional attention, which prevents KV-cache reuse and is expensive for long responses. Block-wise dLLMs~\citep{arriola2025block,cheng2026sdar} partition the response into $B$ consecutive blocks $Y_{1},\ldots,Y_{B}$ of size $K$, where $Y_{b}=(y^{(b-1)K+1},\ldots,y^{bK})$. Attention is bidirectional within a block and causal across blocks, so blocks are generated sequentially while tokens within each block are denoised in parallel. This yields the block-causal factorization
\begin{equation}
    \pi_{\theta}(y\mid x)
    =
    \prod_{b=1}^{B}
    \pi_{\theta}(Y_{b}\mid x,Y_{<b})
    \label{eq:block_factorization}
\end{equation}
which preserves KV-cache compatibility across blocks while retaining intra-block parallelism.

\paragraph{Confidence-driven decoding.}
At each denoising step $s$, the model scores every masked position by its max-confidence $p_{i}=\max_{v\in\mathcal{V}} p_{\theta}(y_{0}^{i}=v\mid y_{s},x)$, where $y_{s}$ is the partially denoised sequence at step $s$, and commits position $i$ only when $p_{i}>\tau$ for a fixed threshold $\tau$. Aggressive thresholds (small $\tau$) commit more tokens per step but inject errors that propagate through later denoising. Conservative thresholds preserve quality but require more refinement steps, eroding the parallelism advantage of dLLMs.

\subsection{GRPO for dLLMs}
\label{sec:prelim_grpo}
For each prompt $x\sim\mathcal{D}$, the old policy samples a group of $G$ responses $\{y_{j}\}_{j=1}^{G}\sim\pi_{\theta_{\mathrm{old}}}(\cdot\mid x)$ with scalar rewards $\{r_{j}\}_{j=1}^{G}$ and the standardized advantage $\hat{A}_{j}=(r_{j}-\operatorname{mean}\{r_{i}\})/\operatorname{std}\{r_{i}\}$. Standard GRPO~\citep{shao2024deepseekmath} defines a token-level importance ratio $\rho_{j}^{k}(\theta)=\pi_{\theta}(y_{j}^{k}\mid x,y_{j}^{<k})/\pi_{\theta_{\mathrm{old}}}(y_{j}^{k}\mid x,y_{j}^{<k})$ and maximizes the per-token clipped surrogate $\frac{1}{G}\sum_{j}\frac{1}{|y_{j}|}\sum_{k}\min\!\bigl(\rho_{j}^{k}\hat{A}_{j},\operatorname{clip}(\rho_{j}^{k},1{-}\epsilon,1{+}\epsilon)\hat{A}_{j}\bigr)$, regularized by $\beta D_{\mathrm{KL}}[\pi_{\theta}\,\|\,\pi_{\mathrm{ref}}]$ toward a reference policy.

For dLLMs, the per-token conditional $\pi_{\theta}(y_{j}^{k}\mid x,y_{j}^{<k})$ is intractable in closed form because the model parameterizes $p_{\theta}(\cdot\mid y_{t},x)$ at masked positions rather than via a left-to-right factorization. Random-masking dLLM-RL methods~\citep{gong2025diffucoder,yang2026mmada} factorize the sequence probability into per-token denoising
terms and approximate the ratio on sampled corruptions $\widetilde{y}_{j}(t)$ of each response at masking level $t\sim \mathrm{U}[0,1]$:
\begin{equation}
    \rho_{j}^{k}(\theta) \;\approx\; \frac{p_{\theta}\bigl(y_{j}^{k}\mid x,\widetilde{y}_{j}(t)\bigr)}{p_{\theta_{\mathrm{old}}}\bigl(y_{j}^{k}\mid x,\widetilde{y}_{j}(t)\bigr)},\quad k\in\mathcal{U}_{j}(t),
    \label{eq:dllm_ratio}
\end{equation}
where $\mathcal{U}_{j}(t)$ denotes the masked positions in $\widetilde{y}_{j}(t)$.

\section{Method}
\label{sec:method}

SLIM-RL separates dLLM RL into two stages: risk-controlled rollout collection, then trace-free low-variance optimization. A $\tau$-budget decoder (Section~\ref{sec:method_unmask}) shapes the rollouts by bounding the cumulative uncertainty committed at each denoising step; a low-variance random-masking objective then updates the policy on them (Sections~\ref{sec:block_obj} and~\ref{sec:var_reduction}), with a mean-preserving decreasing per-block mask schedule (Section~\ref{sec:mask_schedule}) concentrating the gradient signal on the earliest, most-conditioned block.

\subsection{$\tau$-Budget Dynamic Unmasking}
\label{sec:method_unmask}

$\tau$-budget dynamic unmasking is training-free and used for both RL rollout generation and test-time inference. At denoising step $s$, let $\mathcal{M}_s$ denote the currently masked positions in the active block, and let
$p_i = \max_{v\in\mathcal{V}} p_{\theta}(y_0^{i}=v\mid y_s, x)$
be the model's max-confidence at position $i$. The threshold-based candidate
set
\begin{equation}
    \mathcal{A}_s = \{i \in \mathcal{M}_s : p_i > \tau\}
    \label{eq:candidate_set}
\end{equation}
is the set of positions standard dynamic sampling would commit at step $s$. We assign each candidate the confidence-based uncertainty $u_i = 1-p_i$ and define the cumulative step-level uncertainty of a set $\mathcal{S}\subseteq\mathcal{A}_s$ as $U(\mathcal{S})=\sum_{i\in\mathcal{S}} u_i=\sum_{i\in\mathcal{S}}(1-p_i)$. Given a risk-budget schedule $m\ge 1$, we set the step-level budget
$\mathcal{B}_s = m(1-\tau)$. We sort the candidates by
ascending uncertainty
$u_{i_1}\le u_{i_2}\le\cdots\le u_{i_{|\mathcal{A}_s|}}$
and commit the largest prefix of this ordering whose cumulative uncertainty stays within the
budget:
\begin{equation}
\begin{aligned}
    k_s
    &=
    \max\left\{
        k:
        \sum_{r=1}^{k} u_{i_r} \le \mathcal{B}_s
    \right\},\\
    \mathcal{C}_s &= \{i_1,\ldots,i_{k_s}\}.
\end{aligned}
\label{eq:budget_constraint}
\end{equation}

\begin{figure*}[t]
\centering
\includegraphics[width=\textwidth]{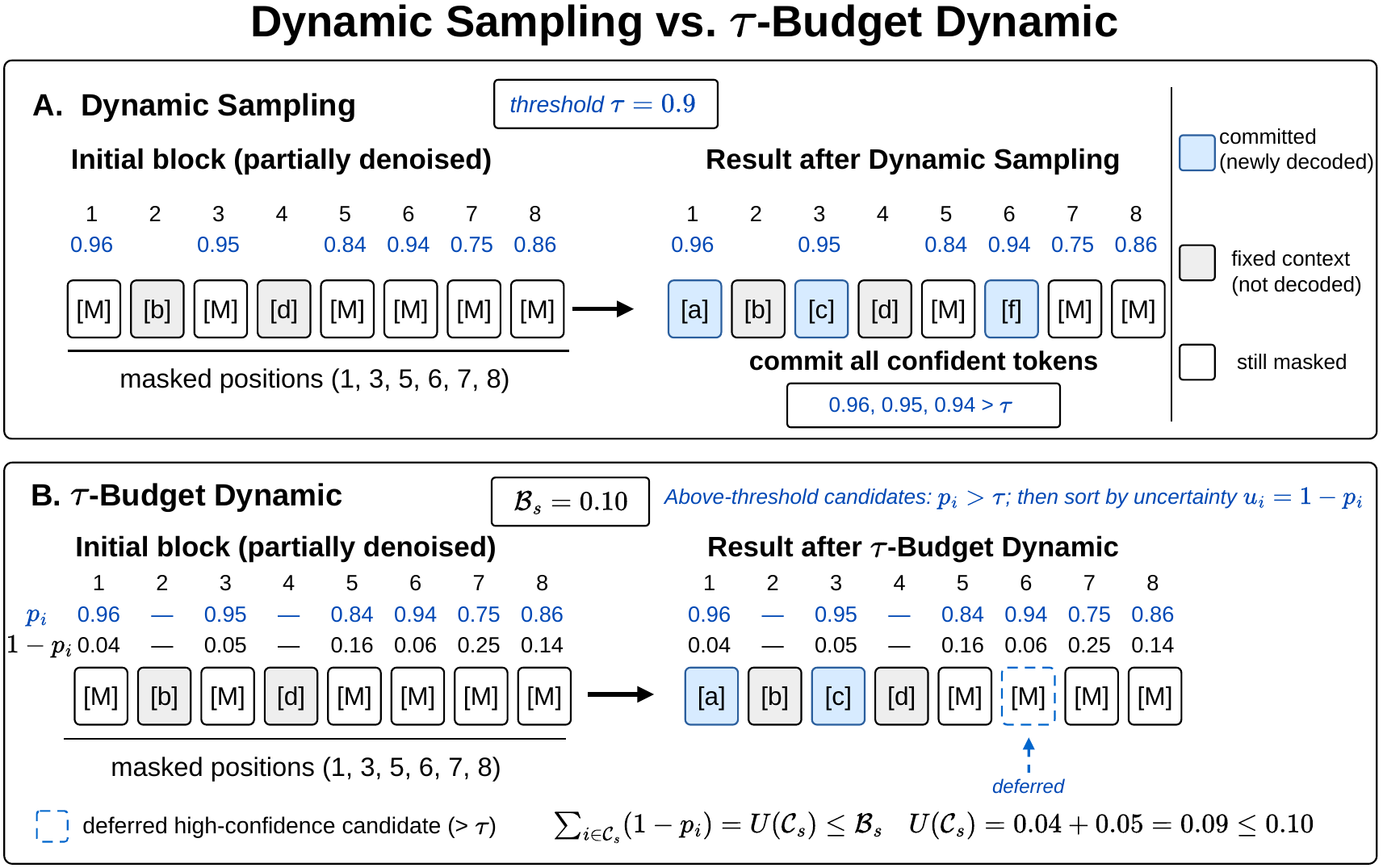}
\caption{\textbf{Dynamic sampling versus $\tau$-budget unmasking at one denoising step.} Both rules start from the same partially denoised block and the same above-threshold candidate set $\mathcal{A}_s$, the three positions whose confidence exceeds $\tau{=}0.9$. \emph{(A)} Dynamic sampling commits every candidate, finalizing all three including the position with confidence $0.94$. \emph{(B)} $\tau$-budget assigns each candidate the uncertainty $u_i{=}1{-}p_i$, sorts ascending, and commits only the largest prefix whose cumulative uncertainty stays within the budget $\mathcal{B}_s{=}m(1{-}\tau){=}0.10$. }
\label{fig:taubudget}
\end{figure*}

The remaining positions stay masked for subsequent denoising steps. If
$\mathcal{A}_s = \emptyset$, we commit the single masked position with the
highest confidence. Figure~\ref{fig:taubudget} contrasts $\tau$-budget with dynamic sampling on a single denoising step, and Algorithm~\ref{alg:tau_budget} gives the full single-pass decoder.
\paragraph{Interpretation: a calibrated wrong-commit budget.}
\label{sec:tau_budget_analysis}
Let $\mathrm{WrongCommit}_s=\sum_{i\in\mathcal{C}_s}\mathbf{1}\{\hat{y}_i\neq y_i^{\ast}\}$ count committed positions that disagree with the correct token. If max-confidence approximated the probability that the argmax is correct, then $\mathbb{E}[\mathrm{WrongCommit}_s]\approx\sum_{i\in\mathcal{C}_s}(1-p_i)=U(\mathcal{C}_s)\le\mathcal{B}_s$, so the budget caps a calibrated proxy for the expected wrong commitments per step. Pointwise thresholding instead commits all of $\mathcal{A}_s$, whose aggregate uncertainty grows with the number of accepted tokens, so many marginally-above-threshold positions can together inject large step-level risk. The calibration assumption holds only approximately, so we treat it as motivation and $\tau$-budget rollouts commit fewer expected-wrong tokens per step than dynamic sampling (Figure~\ref{fig:wrongcommit}).

\subsection{Training Objective: Variance-Reduced Random Masking}
\label{sec:block_obj}
After rollouts are collected, SLIM-RL does not reconstruct the token-level decoding order. The intrinsic weakness of random masking for RL is a high-variance policy gradient. The first source of variance is the granularity of the importance ratio. A token-level ratio attaches a response-level advantage to many independently clipped per-position ratios. As argued by GSPO~\citep{zheng2025group}, such token-level importance weights introduce high-variance training noise that can accumulate with response length. We therefore use a sequence-level ratio, which is computed for the whole response at a given masking level.

For a masking level $t$, let $\widetilde{y}_{j}(t)$ be a randomly masked version of $y_j$, let $\mathcal{U}_{j}(t)$ denote its masked positions, and let $M_{j}(t)=|\mathcal{U}_{j}(t)|$. The normalized sequence denoising log-score is
\begin{equation}
    \ell_{\theta}^{j}(t)
    =
    \frac{1}{M_{j}(t)}
    \sum_{i\in\mathcal{U}_{j}(t)}
    \log
    p_{\theta}
    \left(
        y_j^{i}
        \mid
        x,\widetilde{y}_{j}(t)
    \right),
    \label{eq:block_log_score}
\end{equation}
where $y_j^{i}$ is the clean token at masked position $i$.

The corresponding sequence-level denoising ratio is the geometric mean of the per-position denoising ratios over the masked positions,
\begin{equation}
\rho_{j}^{(t)}(\theta)
=
\exp\!\left(
    \ell_{\theta}^{j}(t)
    -
    \ell_{\theta_{\mathrm{old}}}^{j}(t)
\right)
=
\exp\!\left(
\frac{1}{M_{j}(t)}
\sum_{i\in\mathcal{U}_{j}(t)}
\log
\frac{
    p_{\theta}( y_j^{i} \mid x,\widetilde{y}_{j}(t) )
}{
    p_{\theta_{\mathrm{old}}}( y_j^{i} \mid x,\widetilde{y}_{j}(t) )
}
\right).
\label{eq:block_importance_ratio}
\end{equation}

\subsection{Deterministic Quadrature and Unnormalized Advantage}
\label{sec:var_reduction}

The second source of variance is the masking level $t$, over which the random masking objective is an expectation. Prior dLLM-RL estimates this expectation stochastically, drawing $t$ at random for each update~\citep{gong2025diffucoder, yang2026mmada}, and \citet{rojas2025improving} show that this random time is the dominant source of the estimator's variance. We instead evaluate the expectation deterministically with Gauss--Legendre quadrature over $(0,1)$. At $Q$-point nodes and weights $\{(t_q,\omega_q)\}_{q=1}^{Q}$, $\rho_{j}^{(t_q)}(\theta)$ is the ratio of Eq.~\ref{eq:block_importance_ratio} at $t{=}t_q$, which removes the variance of the random $t$. We also use the unnormalized advantage to remove bias~\citep{liu2025understanding}, and let $A_j$ denote the response-level advantage assigned to $y_j$. 

\begin{equation}
    A_j
    =
    r_j
    -
    \frac{1}{G}
    \sum_{i=1}^{G}
    r_i.
    \label{eq:dr_centered_advantage}
\end{equation}

\paragraph{Final objective.}
The final objective combines the per-quadrature-node sequence ratios $\rho_{j}^{(t_q)}(\theta)$, the unnormalized advantage $A_j$ of Eq.~\ref{eq:dr_centered_advantage}, and a KL regularizer toward the old policy.

\begin{equation}
\mathcal{J}_{\mathrm{SLIM}}(\theta)
=
\mathbb{E}
\Biggl[
\frac{1}{G}
\sum_{j=1}^{G}
\sum_{q=1}^{Q}
\omega_q\,
\psi_{\epsilon}\!\left(\rho_{j}^{(t_q)}(\theta),\,A_j\right)
-
\beta\,
D_{\mathrm{KL}}\!\left(\pi_{\theta}\,\|\,\pi_{\theta_{\mathrm{old}}}\right)
\Biggr].
\label{eq:final_objective}
\end{equation}

We define the clipped policy objective term as $\psi_{\epsilon}(\rho, A)=\min\!\left(\rho A,\,\operatorname{clip}(\rho,1-\epsilon,1+\epsilon)A\right)$. Algorithm~\ref{alg:slim_train} details the training pipeline, from risk-budgeted rollout collection to the quadrature-based policy update.

\subsection{Monotonically Decreasing Per-Block Masking Schedule}
\label{sec:mask_schedule}
Vanilla random-masking RL masks each position independently. We instead apply a monotonically decreasing per-block schedule $p_1\ge p_2\ge\cdots\ge p_B$ across the $B$ blocks $Y_1,\dots,Y_B$ of a response, masking block $Y_b$ at rate $p_b$. Under the block-causal factorization of Eq.~\eqref{eq:block_factorization} the earliest block conditions every later block, so masking it most concentrates the learning signal on the positions the rest of the response depends on, while later blocks stay progressively less masked. The per-block rates average to the global masking level $t$ ($\tfrac1B\sum_b p_b=t$) and the explicit cosine form is given in Appendix~\ref{sec:schedule_formula}. It reaches $32.91$ on MATH500 against $27.00$ for uniform masking (Table~\ref{tab:ablation}).


\section{Experiments}
\label{sec:exp}

\subsection{Setup}
\label{sec:exp_setup}

\paragraph{Data, models, and training.}
Our base model is the block-wise dLLM SDAR-4B-Chat~\citep{cheng2026sdar} with block size 16 and block size 4, SDAR-1.7B-Chat is at block size 4. Following \citet{wang2025revolutionizing} we train on the MATH set~\citep{hendrycks2021measuring} (level 3--5) and PrimeIntellect-verified coding problems~\citep{jaghouar2024intellect1technicalreport}, keeping a separate model per task. On SDAR-4B-Chat this runs for $160$ steps at block size 16 and $100$ steps at block size 4, and the SDAR-1.7B-Chat model at block size 4 trains to convergence. The code model continues from the 4B math model at block size 4 and also trains to convergence. Each RL step samples $128$ prompts with $G{=}8$ responses each at temperature $1.0$. We use $Q{=}3$ Gauss--Legendre nodes per step, AdamW at constant learning rate $1{\times}10^{-6}$, clip $\epsilon{=}0.1$ (the range $[1{-}0.1,1{+}0.1]$), and KL coefficient $\beta{=}0.01$ ($k_3$ estimator), on $8\times$A100 (40\,GB) GPUs across two nodes of four (full configuration in Appendix~\ref{sec:appendix}, Table~\ref{tab:hyperparams}, and the evaluation details in Appendix~\ref{sec:appendix_eval}).

\paragraph{Evaluation.}
We evaluate on two math benchmarks, MATH500~\citep{hendrycks2021measuring} and GSM8K~\citep{cobbe2021training}, and two code benchmarks, MBPP~\citep{austin2021program} and HumanEval~\citep{chen2021evaluating}. Generation uses length $256$, temperature $1.0$, and the two decoders, dynamic sampling and $\tau$-budget at both block size 16 and block size 4. We compare against an autoregressive reference, Qwen2.5-7B-Instruct~\citep{qwen2025qwen25technicalreport}; the full-attention dLLMs LLaDA-8B-Instruct~\citep{nie2026large} and Dream-7B-Instruct~\citep{ye2025dream}; and two RL baselines, random-masking and TraceRL.
\subsection{Main Results}
\label{sec:exp_main}

\begin{table*}[t]
\centering
\small
\setlength{\tabcolsep}{3pt}
\resizebox{\textwidth}{!}{%
\begin{tabular}{lcccccccc}
\toprule
& \multicolumn{2}{c}{MATH500} & \multicolumn{2}{c}{GSM8K} & \multicolumn{2}{c}{MBPP} & \multicolumn{2}{c}{HumanEval} \\
\cmidrule(lr){2-3} \cmidrule(lr){4-5} \cmidrule(lr){6-7} \cmidrule(lr){8-9}
Method & dynamic & $\tau$ & dynamic & $\tau$ & dynamic & $\tau$ & dynamic & $\tau$ \\
\midrule
\multicolumn{9}{l}{\emph{Reference models}} \\
Qwen2.5-7B-Instruct~\citep{qwen2025qwen25technicalreport} & \multicolumn{2}{c}{49.67} & \multicolumn{2}{c}{90.65} & \multicolumn{2}{c}{61.07} & \multicolumn{2}{c}{78.66} \\
LLaDA-8B-Instruct~\citep{nie2026large} & 36.33 & 37.40 & 82.34 & 82.59 & 37.56 & 37.47 & 40.85 & 41.67 \\
Dream-7B-Instruct~\citep{ye2025dream} & 33.40 & 33.47 & 55.52 & 54.41 & 36.93 & 37.33 & 35.57 & 36.18 \\
\midrule
\multicolumn{9}{l}{\emph{From SDAR-4B-Chat, block size 16 (math RL)}} \\
SDAR-4B-Chat~\citep{cheng2026sdar} & 12.18 & 12.07 & 43.54 & 45.14 & 38.80 & \textbf{40.56} & 54.48 & \textbf{56.49} \\
\quad + Random-masking RL~\citep{zhao2026d1} & 12.82 & 13.36 & 47.71 & 48.14 & 38.69 & 40.07 & \textbf{55.22} & 55.76 \\
\quad + TraceRL~\citep{wang2025revolutionizing} & 25.04 & 26.09 & 59.99 & 64.19 & 35.96 & 36.96 & 44.65 & 48.78 \\
\quad + Ours & \textbf{31.36} & \textbf{32.91} & \textbf{71.04} & \textbf{75.03} & \textbf{38.84} & 39.20 & 53.39 & 53.32 \\
\midrule
\multicolumn{9}{l}{\emph{From SDAR-4B-Chat, block size 4 (math RL, then code RL continued from math)}} \\
SDAR-4B-Chat (base) & 10.73 & 10.47 & 62.62 & 62.09 & 50.96 & 51.29 & 66.89 & 66.40 \\
\quad + TraceRL (math RL) & 46.76 & \textbf{46.40} & 88.07 & \textbf{88.32} & 50.71 & 50.64 & 67.89 & 67.01 \\
\quad + Ours (math RL) & \textbf{47.09} & 46.02 & \textbf{88.30} & 87.87 & 51.09 & 49.93 & 68.77 & 67.48 \\
\quad + TraceRL (code RL) & 46.58 & 45.89 & 88.11 & 88.59 & 50.22 & 50.76 & 68.50 & 67.41 \\
\quad + Ours (code RL) & 46.91 & 47.40 & 88.16 & 88.08 & \textbf{54.42} & \textbf{55.73} & \textbf{72.15} & \textbf{73.78} \\
\midrule
\multicolumn{9}{l}{\emph{From SDAR-1.7B-Chat, block size 4 (math RL)}} \\
SDAR-1.7B-Chat (base) & 11.40 & 11.40 & 61.08 & 62.12 & 39.04 & 39.60 & \textbf{51.02} & 50.75 \\
\quad + TraceRL (math RL) & 35.29 & 35.78 & 76.35 & 76.50 & 40.22 & 40.96 & 50.47 & 50.20 \\
\quad + Ours (math RL) & \textbf{36.71} & \textbf{36.47} & \textbf{76.50} & \textbf{77.48} & \textbf{40.64} & \textbf{41.09} & 50.88 & \textbf{53.73} \\
\bottomrule
\end{tabular}
}

\caption{Main results (accuracy, \%). For each benchmark the paired sub-columns dynamic and $\tau$ are two test-time decoders applied to every row: dynamic sampling and our $\tau$-budget decoder. Ours\,=\,SLIM-RL, which collects its training rollouts with the $\tau$-budget decoder; the dynamic and $\tau$ columns are test-time choices, independent of the rollout decoder. Upper group: block size 16 (math RL); lower groups: block size 4 (math RL, then code RL continued from math).}
\label{tab:main_results}
\end{table*}

\paragraph{SLIM-RL substantially outperforms the RL baselines on math.}
On the SDAR-4B-Chat base at block size 16, SLIM-RL reaches $31.36$ on MATH500 and $71.04$ on GSM8K under dynamic sampling, and $32.91$ and $75.03$ under $\tau$-budget, well above TraceRL at $25.04$ and $59.99$ and random-masking RL, which barely moves off the base at $12.82$ and $47.71$. At block size 4 it stays ahead of TraceRL on both math sets under dynamic sampling, and the 4B model surpasses the larger LLaDA-8B and Dream-7B diffusion models, exceeding LLaDA-8B by $10.76\%$ on MATH500 while remaining below the autoregressive Qwen2.5-7B (Table~\ref{tab:main_results}). The advantage also carries to a smaller model, where on SDAR-1.7B at block size 4, SLIM-RL again outperforms TraceRL on MATH500 ($36.71$ vs.\ $35.29$ under dynamic sampling) and matches it on GSM8K, with cross-scale data-efficiency reported in Section~\ref{sec:exp_efficiency}.
\paragraph{SLIM-RL also outperforms TraceRL on code.}
Under code RL trained from the block size 4 math model, SLIM-RL improves over TraceRL by $4.20\%$ on MBPP and $3.65\%$ on HumanEval under dynamic sampling, while preserving math accuracy. Block size 4 parallelism (tokens-per-forward, TPF) and the decode-time block-size change are reported in Appendix~\ref{sec:blocksize_code}.

\subsection{Ablation Study}
\label{sec:exp_training}

Each component contributes as a leave-one-out drop from the full recipe on SDAR-4B-Chat at block size 16, measured on MATH500 under the $\tau$-budget decoder where the full recipe peaks at $32.91$ (Table~\ref{tab:ablation}). The largest drop is the quadrature, where using two Gauss--Legendre quadrature nodes instead of three lowers it to $26.56$. Generating the rollouts with dynamic sampling instead of $\tau$-budget lowers accuracy to $27.09$. The sequence-level importance ratio attains the full-recipe $32.91$, whereas a token-level ratio degrades it to $28.09$. Replacing the monotonically decreasing per-block schedule with uniform random masking lowers accuracy to $27.00$. The $\tau$-budget rollout's benefit does not transfer to TraceRL. Trained on $\tau$-budget rollouts, TraceRL reaches only $22.91$ on MATH500, below the $26.09$ of its usual dynamic sampling rollouts, so the gain is specific to trace-free random masking.

\subsection{Scaling Block Size}
\label{sec:exp_blocksize}

SLIM-RL's advantage over TraceRL grows with the block size. With both methods trained natively at each block size (Table~\ref{tab:blocksize}), the two are on par at block size 4, where TraceRL keeps the exact ($s{=}1$) trajectory at low slicing cost and SLIM-RL matches it on math to within $0.33\%$ on MATH500 and $0.23\%$ on GSM8K under matched decoding. At block size 16 the math margin widens to $6.32\%$ and $11.05\%$. The same widening shows on code benchmarks. TraceRL loses code ability at block size 16, its MBPP falling to $35.96$ and HumanEval to $44.65$ from the base's $38.80$ and $54.48$, while SLIM-RL holds near the base at $38.84$ and $53.39$ (Table~\ref{tab:main_results}). Random-masking RL improves far less than TraceRL and SLIM-RL on math at every block size, while on code it stays near the base.

\subsection{Training Efficiency: Trace-Free Optimization Avoids Trajectory Slicing}
\label{sec:exp_efficiency}

\paragraph{Trace-free optimization attains TraceRL's accuracy at a constant per-rollout cost.}
On SDAR-4B the baseline produces $3.55$ samples per response at block size 16 ($s{=}4$) and $4.00$ at block size 4 ($s{=}1$), both higher than our constant $Q{=}3$; the full trajectory ($s{=}1$) at block size 16 instead costs $12.92$, with $18.5\%$ of responses reaching the full $16$ samples.\footnote{Measured over two passes of MATH500 ($1000$ responses) under the exact TraceRL slicing.} At block size 16, plain random-masking RL is near the base model while SLIM-RL overtakes TraceRL, reaching its best MATH500 accuracy on only $0.46\times$ the training data (Table~\ref{tab:dataeff}, Figure~\ref{fig:efficiency}). At block size 4 and $1.7$B, it reaches TraceRL's best on $0.81\times$ and $0.76\times$ the data, and uses less total data at every scale, $0.85\times$, $0.70\times$, and $0.74\times$. At block size 4 the baseline keeps the full trajectory and both methods are near their best accuracy, where SLIM-RL still stays slightly ahead.

\paragraph{SLIM-RL also trains to a more parallel model.} Beyond accuracy, SLIM-RL commits more tokens-per-forward than TraceRL at matched decoding. At block size 4 it reaches TPF $1.94$ vs.\ $1.72$ on MATH500 under dynamic sampling and leads on every benchmark (Table~\ref{tab:nativek4}), and at block size 16 it sustains the higher TPF throughout training (Figure~\ref{fig:dynamics}). Because the decoder is fixed, this TPF gain comes from the trained model, not from the $\tau$-budget rule, which at $m{=}1$ slightly lowers TPF in exchange for accuracy. The other panels of Figure~\ref{fig:dynamics} report the training dynamics, where generation length shortens for all methods, and the mask ratio for Random-masking and ours stays near $0.5$ while TraceRL's stays near $0.25$.

\paragraph{The $\tau$-budget decoder transfers across architectures.} Because it reads only per-position confidences, the $\tau$-budget decoder transfers training-free across LLaDA, Dream, and SDAR, a drop-in replacement for dynamic sampling (Table~\ref{tab:crossarch}). On the same model its more conservative commits trade a little throughput for accuracy, leaving tokens-per-forward just below dynamic sampling. Dream-7B is the exception, where $\tau$-budget instead raises tokens-per-forward above dynamic sampling on every benchmark except GSM8K, so it is faster while its code accuracy also rises.

\section{Related Work}
\label{sec:related}

\paragraph{Reinforcement learning for dLLMs.}
Existing RL for dLLMs either reconstructs the decoding trajectory or trains on randomly masked corruptions of the completed response. On the trajectory side, TraceRL~\citep{wang2025revolutionizing} optimizes the objective on the exact decoding trace, DCoLT~\citep{huang2026reinforcing} rewards the full denoising trajectory through outcome-based RL, and d2~\citep{wang2026d2improvingreasoningdiffusion} estimates the likelihood of the decoding trajectory. The masking view instead discards the decoding order, with d1~\citep{zhao2026d1} factorizing the policy through a mean-field decomposition, DiffuCoder's coupled-GRPO~\citep{gong2025diffucoder} pairing complementary masking realizations, and MMaDA's UniGRPO~\citep{yang2026mmada} adapting GRPO to masked diffusion. Closest to our objective, \citet{rojas2025improving} replace the random masking level with Gaussian quadrature on the diffusion ELBO, and ESPO~\citep{ou2025principled} reduces the ratio to a single sequence-level weight per response. SLIM-RL stays on the random-masking side and controls its gradient variance by recombining existing components, a sequence-level length-normalized ratio~\citep{zheng2025group}, deterministic quadrature over the masking level~\citep{rojas2025improving}, and an unnormalized advantage~\citep{liu2025understanding}. Both act on the training objective alone, whereas SLIM-RL pairs it with the $\tau$-budget rollout decoder and the mean-preserving per-block mask schedule, controlling rollout commit risk together with optimization variance.

\paragraph{Diffusion language models.}
Diffusion language models~\citep{austin2021structured,gong2022diffuseq,li2022diffusion} differ in architecture and in how they commit tokens at inference. Full-attention dLLMs (LLaDA~\citep{nie2026large,bie2025llada20scalingdiffusionlanguage}, Dream~\citep{ye2025dream}, MMaDA~\citep{yang2026mmada}) scale iterative denoising to the 7--8B range but cannot natively reuse KV caches across denoising steps. Block-wise hybrids (Block Diffusion~\citep{arriola2025block}, SDAR~\citep{cheng2026sdar}) generate autoregressively across blocks while denoising the active block in parallel, restoring KV-cache reuse and serving as the substrate for recent dLLM RL~\citep{wang2025revolutionizing,hu2026lightningrl,zhu2025dirl}. A separate line accelerates inference by adaptively committing tokens at each denoising step. Fast-dLLM~\citep{wu2025fast} and Dimple~\citep{yu2025dimple} commit every position whose confidence exceeds a threshold (dynamic sampling), D2F~\citep{wang2025diffusion} adds discrete diffusion forcing for faster-than-autoregressive inference, and S2D2~\citep{han2026s2d2} adds training-free self-speculation. Dynamic sampling bounds only per-position confidence, so the wrong-commit risk of a step grows with the number of positions committed. The $\tau$-budget decoder (Section~\ref{sec:method_unmask}) instead bounds the cumulative step-level risk directly in a single pass.

\section{Conclusion}

We asked whether reconstructing the exact decoding trajectory is necessary to match TraceRL at equal training cost in diffusion language models, and found that it is not. SLIM-RL is trace-free, pairing a $\tau$-budget rollout decoder that caps a confidence-based proxy for step-level commit risk with a variance-reduced random-masking objective built from sequence-level ratios, deterministic quadrature, and a decreasing per-block mask schedule. On block-causal SDAR-4B, this recipe outperforms TraceRL on both math and code, reaching its best accuracy on $0.46\times$ the training data. Whether exact trajectory reconstruction remains worthwhile at larger blocks or longer responses is left to future work.

\bibliography{colm2026_conference}
\bibliographystyle{colm2026_conference}

\appendix

\section{Additional Experimental Results}
\label{sec:appendix_training_free}

This appendix collects the parallelism/TPF, cross-architecture, block size 4, code-RL, cross-scale, and training-dynamics results referenced from the main text. All numbers are at response length $256$.

\subsection{Ablation and Block-Size Tables}
\label{sec:appendix_ablations}
These tables support the analysis in Sections~\ref{sec:exp_training}--\ref{sec:exp_blocksize}: Table~\ref{tab:ablation} gives the leave-one-out component ablations and Table~\ref{tab:blocksize} reports the block-size dependence of the advantage over TraceRL.

\begin{table}[t]
\centering
\small
\setlength{\tabcolsep}{6pt}
\begin{tabular}{lc}
\toprule
& MATH500 \\
\midrule
Full recipe & \textbf{32.91} \\
\midrule
\multicolumn{2}{l}{\emph{Importance-ratio granularity}} \\
\quad sequence-level (adopted, $=$ full) & \textbf{32.91} \\
\quad token-level & 28.09 \\
\midrule
\multicolumn{2}{l}{\emph{Rollout decoder}} \\
\quad $\tau$-budget ($=$ full) & \textbf{32.91} \\
\quad dynamic & 27.09 \\
\midrule
\multicolumn{2}{l}{\emph{Per-block mask schedule}} \\
\quad monotonic ($=$ full) & \textbf{32.91} \\
\quad random masking & 27.00 \\
\midrule
\multicolumn{2}{l}{\emph{Quadrature points $Q$}} \\
\quad $Q{=}3$ ($=$ full) & \textbf{32.91} \\
\quad $Q{=}2$ & 26.56 \\
\midrule
\multicolumn{2}{l}{\emph{Rollout decoder on TraceRL }} \\
\quad dynamic & 26.09 \\
\quad $\tau$-budget & 22.91 \\
\bottomrule
\end{tabular}
\caption{Leave-one-out ablations of SLIM-RL on MATH500 and ablation to TraceRL.}
\label{tab:ablation}
\end{table}

\begin{table}[t]
\centering
\small
\setlength{\tabcolsep}{4pt}
\begin{tabular}{lcccccc}
\toprule
& \multicolumn{3}{c}{MATH500} & \multicolumn{3}{c}{GSM8K} \\
\cmidrule(lr){2-4}\cmidrule(lr){5-7}
Block size & Random & TraceRL & Ours & Random & TraceRL & Ours \\
\midrule
block size 16 & 12.82 & 25.04 & 31.36 & 47.71 & 59.99 & 71.04 \\
block size 4    & 15.36 & 46.76 & 47.09 & 71.75 & 88.07 & 88.30 \\
\midrule
$\Delta$ at block size 16 & \multicolumn{3}{c}{$+6.32$} & \multicolumn{3}{c}{$+11.05$} \\
$\Delta$ at block size 4  & \multicolumn{3}{c}{$+0.33$} & \multicolumn{3}{c}{$+0.23$} \\
\bottomrule
\end{tabular}
\caption{Block-size dependence of SLIM-RL's advantage over TraceRL (dynamic sampling); block size 4 uses models trained natively at block size 4.}
\label{tab:blocksize}
\end{table}

\subsection{Cross-Architecture Decoder Transfer and Parallelism}
Because $\tau$-budget depends only on per-position confidences, it applies unchanged to full-attention (LLaDA, Dream) and block-causal (SDAR) diffusion models as a training-free, drop-in replacement for dynamic sampling (Table~\ref{tab:crossarch}). Full SLIM-RL is excluded because it already trains with $\tau$-budget, so we include its dynamic-sampling-trained ablation in its place. On parallelism, at the conservative budget $m{=}1$ the $\tau$-budget tokens-per-forward is uniformly slightly below dynamic sampling across LLaDA and every SDAR model in Table~\ref{tab:crossarch}, the small throughput cost it pays for its accuracy gains, so we report it as an accuracy decoder rather than a speed one. Dream-7B at block size 4 is the one architecture where $\tau$-budget speeds up TPF on MATH500, MBPP, and HumanEval, with only GSM8K nearly the same at $2.61$ versus $2.60$; on the two code sets accuracy rises in step, MBPP from $36.93$ to $37.33$ at TPF $4.45$ to $4.59$ and HumanEval from $35.57$ to $36.18$ at TPF $5.46$ to $5.72$. For Dream the decoder is both faster and more accurate on code, while on MATH500 accuracy is nearly unchanged and on GSM8K it drops slightly. 
\begin{table*}[t]
\centering
\small
\setlength{\tabcolsep}{4pt}
\caption{Training-free $\tau$-budget decoding across architectures.}
\label{tab:crossarch}
\begin{tabular}{llccccccccc}
\toprule
& & & \multicolumn{2}{c}{MATH500} & \multicolumn{2}{c}{GSM8K} & \multicolumn{2}{c}{MBPP} & \multicolumn{2}{c}{HumanEval} \\
\cmidrule(lr){4-5}\cmidrule(lr){6-7}\cmidrule(lr){8-9}\cmidrule(lr){10-11}
Model / ckpt & Decoder & $K$ & acc & TPF & acc & TPF & acc & TPF & acc & TPF \\
\midrule
\multicolumn{11}{l}{\emph{Full-attention dLLMs }} \\
LLaDA-8B-Instruct & dynamic           & 32 & 36.33 & 4.07 & 82.34 & 3.20 & 37.56 & 6.34 & 40.85 & 5.08 \\
                  & $\tau$-budget & 32 & 37.40 & 3.79 & 82.59 & 2.86 & 37.47 & 5.69 & 41.67 & 4.66 \\
Dream-7B-Instruct & dynamic           & 4  & 33.40 & 2.19 & 55.52 & 2.61 & 36.93 & 4.45 & 35.57 & 5.46 \\
                  & $\tau$-budget & 4  & 33.47 & 2.21 & 54.41 & 2.60 & 37.33 & 4.59 & 36.18 & 5.72 \\
\midrule
\multicolumn{11}{l}{\emph{Block-causal SDAR-4B }} \\
SDAR-4B-Chat (base) & dynamic           & 16 & 12.18 & 2.07 & 43.54 & 1.98 & 38.80 & 1.60 & 54.48 & 1.75 \\
                    & $\tau$-budget & 16 & 12.07 & 1.95 & 45.14 & 1.87 & 40.56 & 1.47 & 56.49 & 1.60 \\
                    & dynamic           & 4  & 10.73 & 2.09 & 62.62 & 1.92 & 50.96 & 1.55 & 66.89 & 1.68 \\
                    & $\tau$-budget & 4  & 10.47 & 2.03 & 62.09 & 1.85 & 51.29 & 1.48 & 66.40 & 1.59 \\
\;+ Random-masking RL & dynamic           & 16 & 12.82 & 2.09  & 47.71 & 2.08 & 38.69 & 1.61 & 55.22 & 1.76 \\
                    & $\tau$-budget & 16 & 13.36 & 1.97 & 48.14 & 1.94 & 40.07 & 1.48 & 55.76 & 1.64 \\
\;+ TraceRL         & dynamic           & 16 & 25.04 & 1.80  & 59.99 & 2.05 & 35.96 & 1.61 & 44.65 & 1.76 \\
                    & $\tau$-budget & 16 & 26.09 & 1.71 & 64.19 & 1.93 & 36.96 & 1.49 & 48.78 & 1.63 \\
                    & dynamic           & 4  & 46.76 & 1.72 & 88.07 & 2.02 & 50.71 & 1.50 & 67.89 & 1.64 \\
                    & $\tau$-budget & 4  & 46.40 & 1.66 & 88.32 & 1.95 & 50.64 & 1.42 & 67.01 & 1.56 \\
\;+ SLIM-RL (w/o $\tau$) & dynamic           & 16 & 26.29 & 1.97  & 67.73 & 2.30 & 37.56 & 1.66 & 52.91 & 1.90 \\
                    & $\tau$-budget & 16 & 27.09 & 1.85 & 70.38 & 2.13 & 38.80 & 1.53 & 54.81 & 1.73 \\
                    & dynamic           & 4  & 47.22 & 1.95 & 88.07 & 2.08 & 51.18 & 1.56 & 68.70 & 1.73 \\
                    & $\tau$-budget & 4  & 46.84 & 1.85 & 88.07 & 2.01 & 50.78 & 1.49 & 68.09 & 1.64 \\
\bottomrule
\end{tabular}
\par\vspace{2pt}
\end{table*}

\subsection{Block Size 4 and Code RL}
\label{sec:blocksize_code}
Table~\ref{tab:nativek4} reports the full block size 4 results: the two methods reach comparable accuracy, with SLIM-RL slightly ahead, and replacing dynamic sampling with $\tau$-budget decoding leaves accuracy nearly unchanged. The code-RL results, with code RL continued from the block size 4 math-RL model, are reported in the main-text Table~\ref{tab:main_results} (lower group).
\begin{table*}[t]
\centering
\small
\setlength{\tabcolsep}{4pt}
\caption{Native block size 4 evaluation of the SDAR-4B family (both methods trained and decoded at block size 4).}
\label{tab:nativek4}
\begin{tabular}{llccccccccc}
\toprule
& & & \multicolumn{2}{c}{MATH500} & \multicolumn{2}{c}{GSM8K} & \multicolumn{2}{c}{MBPP} & \multicolumn{2}{c}{HumanEval} \\
\cmidrule(lr){4-5}\cmidrule(lr){6-7}\cmidrule(lr){8-9}\cmidrule(lr){10-11}
Method & Decoder & $K$ & acc & TPF & acc & TPF & acc & TPF & acc & TPF \\
\midrule
SLIM-RL (ours)       & dynamic            & 4 & 47.09 & 1.94 & 88.30 & 2.13 & 51.09 & 1.56 & 68.77 & 1.72 \\
                     & $\tau$-budget  & 4 & 46.02 & 1.87 & 87.87 & 2.06 & 49.93 & 1.49 & 67.48 & 1.65 \\
TraceRL (baseline)   & dynamic            & 4 & 46.76 & 1.72 & 88.07 & 2.02 & 50.71 & 1.50 & 67.89 & 1.64 \\
                     & $\tau$-budget  & 4 & 46.40 & 1.66 & 88.32 & 1.95 & 50.64 & 1.42 & 67.01 & 1.56 \\
\bottomrule
\end{tabular}
\end{table*}

\subsection{Training Dynamics and Cross-Scale}
\begin{figure}[t]
\centering
\includegraphics[width=0.62\linewidth]{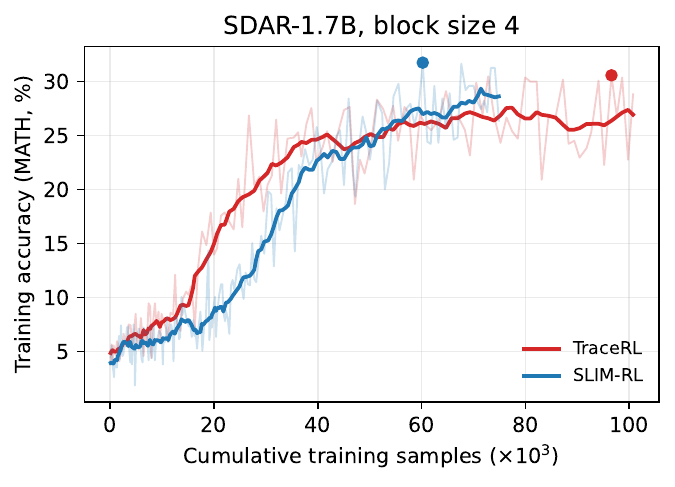}
\caption{Cross-scale training dynamics at $1.7$B and block size 4. SLIM-RL (blue) overtakes TraceRL (red) on training accuracy over cumulative data, reaching TraceRL's best MATH500 accuracy on $0.76\times$ the training samples (Table~\ref{tab:dataeff}). }
\label{fig:eff1p7b}
\end{figure}

\begin{figure*}[t]
\centering
\includegraphics[width=0.92\textwidth]{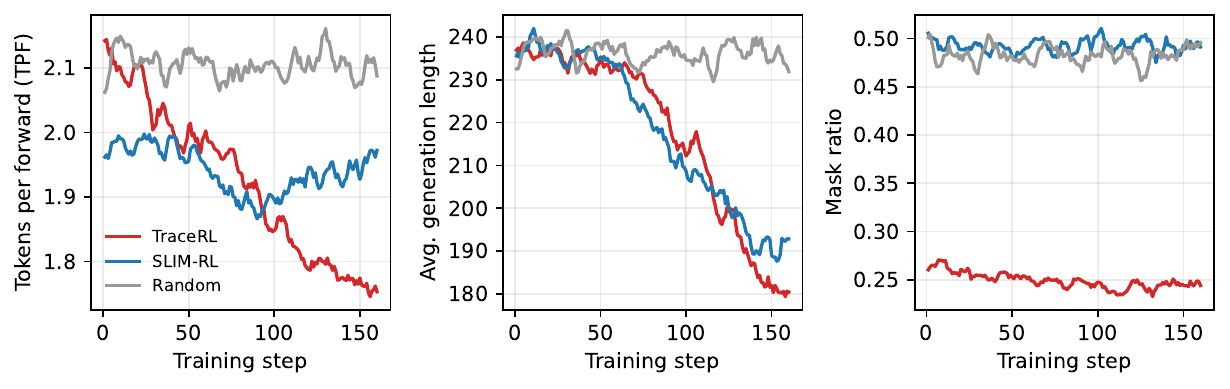}
\caption{Training dynamics at block size 16. \emph{Left.} SLIM-RL maintains a stable tokens-per-forward (TPF) while TraceRL's declines, so SLIM-RL ends higher. \emph{Middle.} Generation length shortens over training for all methods. \emph{Right.} Mask ratio: SLIM-RL and random masking hold near $0.5$ while TraceRL stays near $0.25$.}
\label{fig:dynamics}
\end{figure*}

At $1.7$B scale (Figure~\ref{fig:eff1p7b}, Table~\ref{tab:dataeff}), SLIM-RL stays ahead of TraceRL on MATH500 ($36.71$ vs.\ $35.29$) and reaches TraceRL's best accuracy on $0.76\times$ the training data, a data-efficiency gain that holds at a second model scale. Figure~\ref{fig:dynamics} shows training dynamics: SLIM-RL maintains a stable tokens-per-forward while TraceRL's declines, so SLIM-RL ends higher, and generation length shortens for all methods.
\begin{figure}[t]
\centering
\includegraphics[width=\linewidth]{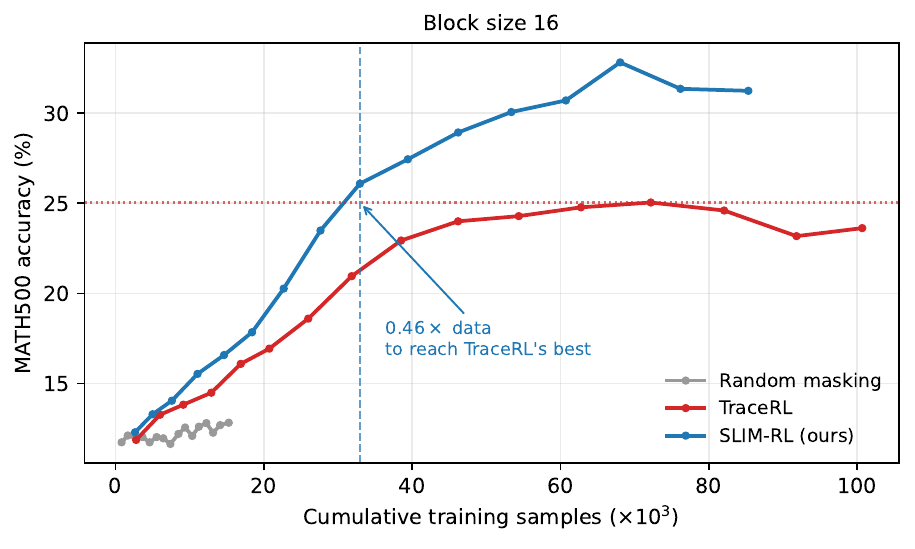}
\caption{MATH500 accuracy versus cumulative training samples at block size 16. Random masking (gray) is near the base model; SLIM-RL (blue) overtakes TraceRL (red), reaching TraceRL's best accuracy on $0.46\times$ the training samples.}
\label{fig:efficiency}
\end{figure}

\begin{table}[t]
\centering
\small
\setlength{\tabcolsep}{6pt}
\begin{tabular}{lcc}
\toprule
Setting & Data to reach ($\times$) & Total-run data ($\times$) \\
\midrule
TraceRL (baseline) & $1.00$ & $1.00$ \\
\midrule
SLIM-RL, SDAR-4B block 16  & $\mathbf{0.46}$ & $0.85$ \\
SLIM-RL, SDAR-4B block 4   & $0.81$ & $\mathbf{0.70}$ \\
SLIM-RL, SDAR-1.7B block 4 & $0.76$ & $0.74$ \\
\bottomrule
\end{tabular}
\caption{Training cost of SLIM-RL relative to the TraceRL baseline (TraceRL $=1.00\times$; $<1$ is cheaper). SLIM-RL needs less data to reach TraceRL's best MATH500 accuracy, and less total-run data, at every scale.}
\label{tab:dataeff}
\end{table}

\begin{figure}[t]
\centering
\includegraphics[width=\linewidth]{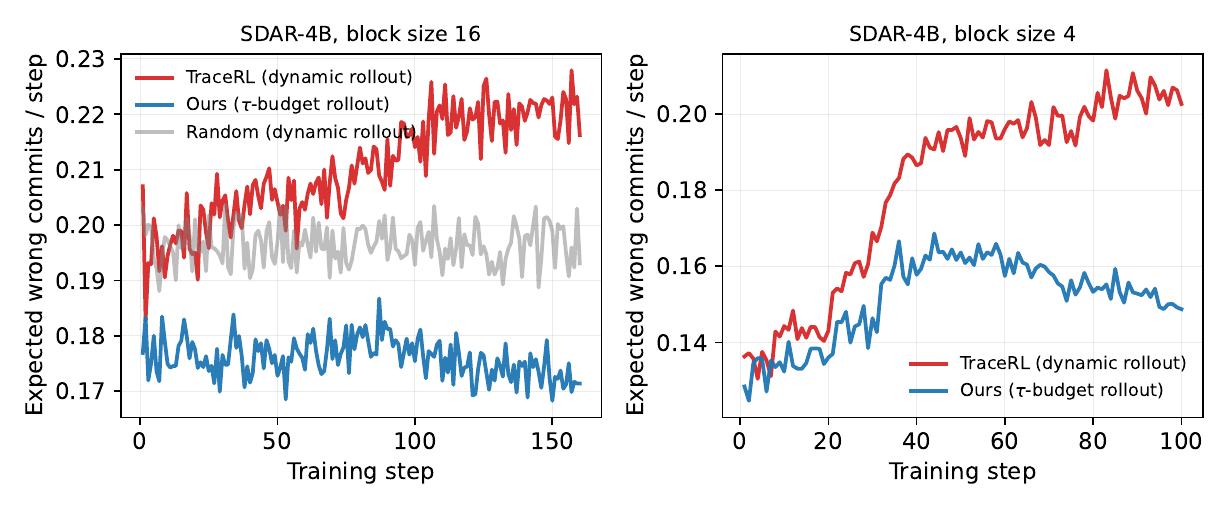}
\caption{Expected wrong commitments per step, $\sum_i(1-p_i)$ over committed positions. The $\tau$-budget rollouts (SLIM-RL, blue) commit fewer than TraceRL's dynamic sampling rollouts, $0.17$ vs.\ $0.22$ at block size 16 and $0.15$ vs.\ $0.21$ at block size 4.}
\label{fig:wrongcommit}
\end{figure}

\subsection{Evaluation Details}
\label{sec:appendix_eval}

We report the average accuracy over $n$ sampled responses per task. For the SDAR family we set $n{=}9$ on MATH500, $n{=}3$ on GSM8K, and $n{=}9$ on both MBPP and HumanEval. A coding response counts as correct only when it passes every functional test. Each diffusion model is evaluated under both decoders, dynamic sampling and the $\tau$-budget decoder of Section~\ref{sec:method_unmask}.

For the SDAR-4B-Chat and SDAR-1.7B-Chat models, we keep the pretrained block-diffusion decoding, a response length of $256$, temperature $1.0$, top-$p{=}1.0$, and top-$k{=}0$. Dynamic sampling unmasks every position above the confidence threshold $\tau{=}0.9$, and the $\tau$-budget decoder reuses that $\tau$ with budget multiplier $m{=}1$. The 4B models are evaluated at block size $16$ and block size $4$, and the 1.7B model at block size $4$.

For the full-attention baselines, we use temperature $0.1$ and sample $n{=}3$ responses per task across all four benchmarks. LLaDA-8B-Instruct decodes at block size $32$ and Dream-7B-Instruct at block size $4$, both over a length-$256$ response. Both baselines run dynamic sampling and the $\tau$-budget decoder at threshold $\tau{=}0.95$ with $m{=}1$.

Qwen2.5-7B-Instruct is the autoregressive reference. We sample $n{=}3$ responses at temperature $0.1$, and raise the generation length to $512$. 

\section{Algorithm Pipeline}
\label{sec:appendix}

Algorithm~\ref{alg:slim_train} details the SLIM-RL training loop. Each outer step samples $G$ rollouts via $\tau$-budget dynamic unmasking (Algorithm~\ref{alg:tau_budget}), scores them with a verifiable reward $r(\cdot)$, forms the unnormalized advantage $A_j$ (Eq.~\ref{eq:dr_centered_advantage}), evaluates the sequence-level ratio $\rho_{j}^{(t_q)}(\theta)$ (Eq.~\ref{eq:block_importance_ratio}) at $Q$ Gauss--Legendre quadrature nodes per response, and updates $\theta$ on the clipped objective (Eq.~\ref{eq:final_objective}).

\begin{algorithm}[t]
\small
\raggedright
\caption{SLIM-RL Training}
\label{alg:slim_train}
\begin{algorithmic}[1]
\State \textbf{Input:}
\State \quad 1) Prompt set $\mathcal D$; reward $r(x,y)$; policy $\pi_\theta$; old policy $\pi_{\theta_{\mathrm{old}}}$.
\State \quad 2) Outer steps $T$; rollouts $G$; update epochs $E$; clip range $\epsilon$; KL weight $\beta$; learning rate $\eta$.
\State \quad 3) Block size $K$; confidence threshold $\tau$; risk-budget schedule $m$; quadrature nodes $Q$.
\State Initialize $\theta$ and Gauss--Legendre nodes/weights $\{(t_q,\omega_q)\}_{q=1}^{Q}$ on $(0,1)$.

\For{$n=1$ to $T$}
    \State $\pi_{\theta_{\mathrm{old}}}\gets \pi_\theta$; \quad $\mathcal D_{\rm grp}\gets\emptyset$
    \Statex \textit{Rollout collection with risk-budgeted decoding}
    \For{prompt minibatch $\mathcal X\sim\mathcal D$}
        \ForAll{$x\in\mathcal X$}
            \State Sample $\{y_{j}\}_{j=1}^{G}$ by
            $\textsc{TauBudgetDecode}(\pi_{\theta_{\mathrm{old}}},x,K,m,\tau)$.
            \State Evaluate $r_{j}\gets r(x,y_{j})$ for $j=1,\ldots,G$.
            \State Compute $A_{j}\gets r_{j}-\frac{1}{G}\sum_{i=1}^{G}r_{i}$.
            \State $\mathcal D_{\rm grp}\gets\mathcal D_{\rm grp}
            \cup\{(x,\{y_{j}\}_{j=1}^{G},\{A_{j}\}_{j=1}^{G})\}$.
        \EndFor
    \EndFor

    \Statex \textit{Policy optimization with sequence-level ratios}
    \For{$e=1$ to $E$}
        \State Sample minibatch $\mathcal G\subset\mathcal D_{\rm grp}$.
        \State Build masked corruptions $\widetilde y_{j}(t_q)$
        for all $(x,y_{j})\in\mathcal G$ and nodes $q$, with per-block rates from the decreasing schedule (Section~\ref{sec:mask_schedule}).
        \State Compute sequence log-scores
        $\ell_{\theta}^{j,q}$ and $\ell_{\theta_{\mathrm{old}}}^{j,q}$ by Eq.~\ref{eq:block_log_score}.
        \State Compute sequence ratios
        $\rho_{j}^{(t_q)}(\theta)$ by Eq.~\ref{eq:block_importance_ratio}.
        \State Update $\theta\gets \theta+\eta\nabla_\theta\mathcal J_{\mathrm{SLIM}}(\theta)$
        using Eq.~\ref{eq:final_objective}.
    \EndFor
\EndFor

\Ensure Trained policy $\pi_\theta$.
\end{algorithmic}
\end{algorithm}

\begin{algorithm}[t]
\small
\caption{\textsc{TauBudgetDecode} ($\tau$-Budget Dynamic Unmasking)}
\label{alg:tau_budget}
\begin{algorithmic}[1]
\State \textbf{Input:}
\State \quad 1) Policy $\pi_\theta$; prompt $x$.
\State \quad 2) Block size $K$; risk-budget schedule $m$; confidence threshold $\tau$.
\State \textbf{Initialize:} $y \gets x \mathbin\Vert [\mathrm{MASK}]^{L}$; block index $b \gets 1$.
\While{some block remains masked}
  \State $s \gets 1$
  \While{block $b$ has masked positions}
    \State For masked $i$ in block $b$: $p_i \gets \max_{v\in\mathcal{V}} p_\theta(y_0^i = v \mid y, x)$.
    \State Candidate set $\mathcal{A}_s \gets \{i : p_i > \tau\}$. \quad // Eq.~\ref{eq:candidate_set}
    \If{$\mathcal{A}_s = \emptyset$}
      \State Commit single $\arg\max_i p_i$ to break tie; \textbf{continue}.
    \EndIf
    \State Uncertainties $u_i \gets 1-p_i$ for $i\in\mathcal{A}_s$; sort $u_{i_1}\le u_{i_2}\le\cdots\le u_{i_{|\mathcal{A}_s|}}$.
    \State Step budget $\mathcal{B}_s \gets m(1-\tau)$.
    \State $k_s \gets \max\{k : \sum_{r=1}^{k} u_{i_r} \le \mathcal{B}_s\}$;\quad committed set $\mathcal{C}_s \gets \{i_1,\ldots,i_{k_s}\}$. \quad // Eq.~\ref{eq:budget_constraint}
    \State For $i\in\mathcal{C}_s$: $\hat y_i \gets \arg\max_{v\in\mathcal{V}} p_\theta(y_0^i = v \mid y, x)$;\quad update $y$.
    \State $s \gets s + 1$
  \EndWhile
  \State $b \gets b + 1$ \quad // move to next block
\EndWhile
\State \textbf{Output:} decoded response $y$.
\end{algorithmic}
\end{algorithm}

\begin{table}[t]
\centering
\small
\setlength{\tabcolsep}{6pt}
\caption{SLIM-RL hyperparameters for the SDAR-4B-Chat math-RL runs.}
\label{tab:hyperparams}
\begin{tabular}{ll}
\toprule
Hyperparameter & Value \\
\midrule
\multicolumn{2}{l}{\textit{Rollout / $\tau$-budget decoding}} \\
Block size $K$ & $16$ \\
Confidence threshold $\tau$ & $0.90$ \\
Risk-budget schedule $m$ & $1$ (constant) \\
\quad effective step budget $\mathcal{B}_s=m(1-\tau)$ & $0.10$ \\
Response length $L$ & $256$ \\
Denoising steps per block & $16$ \\
Sampling temperature & $1.0$ \\
Nucleus / top-$k$ & top-$p\,{=}\,1.0$, top-$k\,{=}\,0$ \\
Tasks per step & $128$ \\
Responses per task (group size $G$) & $8$ \\
\midrule
\multicolumn{2}{l}{\textit{Masking schedule (decreasing per block)}} \\
Quadrature nodes $Q$ & $3$ \\
Quadrature rule & Gauss--Legendre on $(0,1)$ \\
\quad nodes $\{t_q\}$ & $\{0.1127,\,0.5,\,0.8873\}$ \\
\quad weights $\{\omega_q\}$ (sum to $1$) & $\{0.2778,\,0.4444,\,0.2778\}$ \\
Per-block mask-rate spread $\delta$ & $0.2$ \\
\midrule
\multicolumn{2}{l}{\textit{Optimization (SLIM-RL update)}} \\
Clip range (low / high) & $[1{-}0.1,\,1{+}0.1]$ \\
KL coefficient $\beta$ & $0.01$ \\
KL estimator & $k_3$ \\
Optimizer & AdamW \\
Learning rate (constant) $\eta$ & $1{\times}10^{-6}$ \\
AdamW $(\beta_1,\beta_2,\text{wd},\varepsilon)$ & $(0.9,\,0.999,\,0,\,1{\times}10^{-8})$ \\
LR schedule / warmup & cosine, $0$ warmup, $\min$-scale $1.0$ \\
Max gradient norm & $1.0$ \\
Mixed precision & bf16 (TF32 enabled) \\
Outer RL steps $T$ & $160$ \\
Update epochs per step $E$ & $1$ \\
Random seed & $10086$ \\
Hardware & $8\times$A100 40\,GB, 2 nodes $\times$ 4 \\
\bottomrule
\end{tabular}
\end{table}

\subsection{Per-Block Mask Schedule}
\label{sec:schedule_formula}
The monotonically decreasing per-block schedule (Section~\ref{sec:mask_schedule}) is a cosine schedule centered at the masking level $t$:
\begin{equation}
p_b = t + \tfrac{\delta}{2}\cos(\pi \xi_b),\qquad \xi_b=\frac{b-1}{B-1},
\end{equation}
where $\xi_b$ is the normalized position of block $b$ among the $B$ blocks of a response. The rate decreases from $t+\tfrac{\delta}{2}$ at the first block to $t-\tfrac{\delta}{2}$ at the last. Since $\cos(\pi \xi_b)$ is antisymmetric about $\xi_b=\tfrac12$, the per-block rates average to $t$ exactly ($\tfrac1B\sum_b p_b=t$), so the schedule only reallocates a fixed expected mask count across blocks. The spread is bounded by $\delta\le 2\min(t,1-t)$ to keep every $p_b\in[0,1]$; we set $\delta=0.2$ and apply the schedule at each Gauss--Legendre quadrature node $t$.

\end{document}